\documentclass[conference]{IEEEtran}
\IEEEoverridecommandlockouts
\usepackage{cite}
\usepackage{amsmath,amssymb,amsfonts}
\usepackage{algorithmic, algorithm}
\usepackage{graphicx}
\usepackage{textcomp}
\usepackage{xcolor}
\usepackage{multirow}
\def\BibTeX{{\rm B\kern-.05em{\sc i\kern-.025em b}\kern-.08em
    T\kern-.1667em\lower.7ex\hbox{E}\kern-.125emX}}
\begin{document}

\title{IR-UWB Radar-based Situational Awareness System for Smartphone-Distracted Pedestrians
}

%\title{UWB-assisted Situational Awareness System for Distracted Pedestrians}

\author{\IEEEauthorblockN{Jamsheed Manja Ppallan, Ruchi Pandey, Yellappa Damam, Vijay Narayan Tiwari, \\Karthikeyan Arunachalam and Antariksha Ray}
\IEEEauthorblockA{\textit{Connectivity R\&D Group, Samsung R\&D Institute India-Bangalore (SRI-B), India}\\
\{jamsheed.mp, ruchi.1, yellappa.d, vijay.nt, karthikeya.a, antariksha.r\}@samsung.com}
 \\[-3.0ex]
}

\maketitle

\begin{abstract}
With the widespread adoption of smartphones, ensuring pedestrian safety on roads has become a critical concern due to smartphone distraction. This paper proposes a novel and real-time assistance system called UWB-assisted Safe Walk (UASW) for obstacle detection and warns users about real-time situations. The proposed method leverages Impulse Radio Ultra-Wideband (IR-UWB) radar embedded in the smartphone, which provides excellent range resolution and high noise resilience using short pulses. We implemented UASW specifically for Android smartphones with IR-UWB connectivity. The framework uses complex Channel Impulse Response (CIR) data to integrate rule-based obstacle detection with artificial neural network (ANN) based obstacle classification. The performance of the proposed UASW system is analyzed using real-time collected data. The results show that the proposed system achieves an obstacle detection accuracy of up to 97\% and obstacle classification accuracy of up to 95\% with an inference delay of 26.8 ms. The results highlight the effectiveness of UASW in assisting smartphone-distracted pedestrians and improving their situational awareness.
\end{abstract}
\vspace{-0.1cm}

\section{Introduction}
In 2022, the United States witnessed over 6,000 pedestrian fatalities resulting from traffic accidents, and nearly one-third of these tragic incidents were linked to distracted walking \cite{study1,study2,study3}. All the victims were either engrossed in smartphone activities or listening to music. The widespread use of smartphones has become a significant source of distraction for pedestrians, resulting in a rise in accidents and injuries \cite{study4}. Therefore, a smartphone-based solution is needed to alert users of obstacles and be easily accessible to the target audience. While some smartphone features offer basic reminders to be mindful of one's surroundings, they do not analyze the environment or provide real-time alerts, making a sophisticated solution essential for ensuring pedestrian safety \cite{mobtech_book,general_sas}.

%Many researchers demonstrate the potential of radar technology in detecting and addressing distractions  \cite{choi2016multi,ped,nakamura2018performance}. A method based on Bluetooth Low Energy (BLE) beacons was proposed in \cite{bledw} to locate distracted smartphone users. However, this approach necessitates the installation of BLE beacons in each location, restricting its scalability. Despite advancements in image processing and ultrasound-based assistance systems \cite{cameraultrasound,vision1,vision2}, these solutions often lack user convenience. Current approaches typically involve using a smartphone's camera, a Bluetooth-based application, and an ultrasonic sensor \cite{odsosds,sbodvi}, or a wearable device \cite{sodcs,stick_visimp2020}. However, these solutions add additional costs and increase the overall form factor compared to a handheld smartphone. UASW overcomes these limitations by directly integrating the UWB radar chip into the smartphone\cite{apple}.
In the realm of smartphone distraction, various approaches have been explored that include a smartphone's camera \cite{vision1,vision2}, a Bluetooth-based application \cite{bledw}, or wearable devices \cite{sodcs,stick_visimp2020}. The Android-based solution, Google Heads Up\cite{headsup}, was examined for its ability to detect distracted walking, although it focuses solely on user activity and lacks awareness of the surroundings. An approach based on Bluetooth Low Energy (BLE) beacons was proposed in \cite{bledw} to locate distracted smartphone users. However, this method requires the installation of BLE beacons in each location, which can limit its scalability. As described in \cite{vision1} and \cite{vision2}, vision-based systems have shown to be promising but often raise concerns about user privacy and may lack convenience. Furthermore, these solutions typically introduce additional costs and increase the overall form factor compared to a handheld smartphone. %Many researchers have demonstrated the potential of IR-UWB radar technology in detecting human activities and addressing distractions \cite{choi2016multi, ped, nakamura2018performance}. These works have laid the foundation for leveraging IR-UWB radar for obstacle detection and classification.  

The availability of modern smartphones equipped with built-in UWB connectivity presents new opportunities for innovative solutions in this domain \cite{apple, android-uwb}. IR radar offers distinct smartphone advantages, such as low power consumption, finer range resolution, and detecting targets nearby \cite{nakamura2018performance,choi2016multi,mobtech_book}. The short duration of the transmitted waveform in IR-UWB radar (typically in nanoseconds) allows the signal energy to span a wide radio frequency (RF) bandwidth, resulting in immunity to interference from neighboring devices, rendering it highly suitable for obstacle detection\cite{choi2016multi,ped, nakamura2018performance}. These advantages have laid the foundation for leveraging IR-UWB radar for obstacle detection and classification. 

In this paper, we introduce an inventive Situational Awareness System (SAS) that capitalizes on the capabilities of IR-UWB radar technology and the widespread use of smartphones. The proposed UASW directly utilizes the IR-UWB radar chip in the smartphone to overcome the limitations of BLE and vision-based approaches, offering an efficient and privacy-aware solution for distraction detection and pedestrian safety. The proposed application offers the ability to proactively identify potential hazards and deliver timely alerts, thus significantly improving pedestrian safety. The objective of this research is not to encourage or normalize distracted walking but rather to establish an additional layer of protection during walking activities. The system achieves precise hazard detection and alert generation by analyzing radar data and implementing machine learning techniques. The key contributions of this paper can be summarized as follows:
%By leveraging the capabilities of IR-UWB radar and the widespread usage of smartphones, this paper proposes an innovative Situational Awareness System (SAS) to enhance pedestrian situational awareness and promote walk safety significantly. The proposed application can effectively detect potential hazards and provide timely alerts, thereby improving pedestrian safety and awareness. The motivation of this paper is not to normalize or promote distracted walking; instead, it proposes to build an additional layer of protection while walking. Analyzing radar data to comprehend the surroundings and applying machine learning techniques facilitate accurate hazard detection and alert generation. The main contributions of this paper are as follows:
\begin{itemize}
	\item UASW represents the implementation of a radar-based obstacle detection solution for smartphones specifically designed to alert distracted pedestrians.
 \vspace{0.05cm}
	\item We propose an efficient methodology to estimate the presence of obstacles and provide terrain information at a fixed distance from the target based on the smartphone's orientation.
  \vspace{0.05cm}
	\item The proposed UASW overcomes the limitations of Bluetooth, Wi-Fi, and other wireless connectivity-based ranging methods by offering superior positioning accuracy while maintaining low power consumption.
 \vspace{0.05cm}	
 \item The proposed UASW significantly reduces overhead costs for end users compared to existing alternatives and provides a simpler user experience by integrating the solution directly onto the smartphone.
\end{itemize}

The rest of the paper is outlined as follows: Section \ref{sec:sigmod} describes the signal model, system architecture, and operational overview. Section \ref{sec:methodology} explains the rule-based approach for obstacle detection followed by the ML-based approach for obstacle classification. The performance analysis and results of the proposed system are discussed in Section \ref{sec:perfm_analysis} and \ref{sec:results} followed by the conclusion in Section \ref{sec:conclusion}.

\section{Proposed UASW System}
\label{sec:sigmod}
%The math and obstacle detection part will come here. Some description of how UWB radar works can also be given.
The proposed UASW is a smartphone-based system that utilizes low-power IR-UWB radar technology. UASW operates by transmitting precise temporal pulses, receiving the reflected pulses, and deriving the CIR for further processing.
\begin{figure}[h]
	\centering
	\includegraphics[scale=0.72]{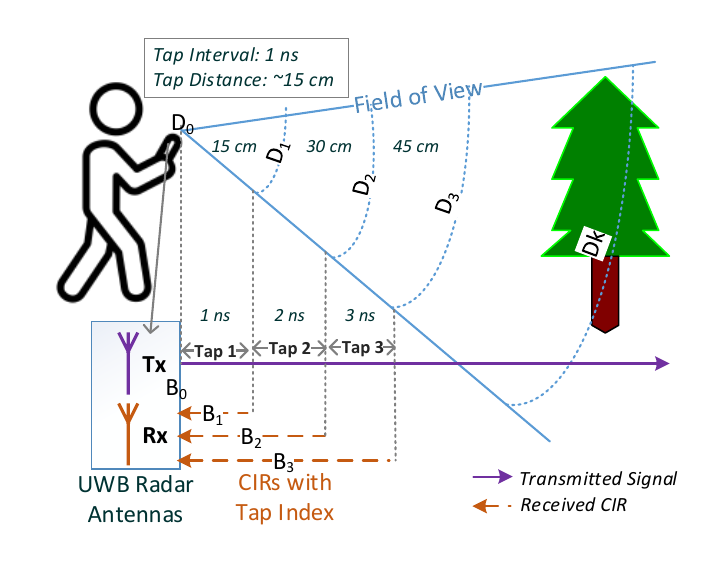}
 \vspace{-0.3cm}
	\caption{IR-UWB radar-enabled smartphones assisting distracted pedestrians.}
	\label{fig:fig_intro}
\end{figure}
\subsection{Signal Model}
Let $x[k]$ be the transmitted short-duration impulse emitted by IR-UWB radar which is being delayed and scaled from $J$ different path is given by \cite{choi2016multi,choi2017people}
\begin{equation}
    s[k] = \sum_{j=1}^{J} a_j x(k-\tau_j) + w[k],
\end{equation}
where $s[k]$ is the CIR captured by UASW during the radar session to facilitate obstacle detection and classification. $w[k]$ represents the noise and $k$ is the digitized sample index.
 %\textbf{check: (transmitted at every 9 ms) - \textcolor{blue}{Transmitted based on the PRF configured. If 200 Hz, at every 5 ms}} signal which gets reflected as $y(t)$ given by
Figure \ref{fig:fig_intro} shows a real-time alert system for smartphone-distracted pedestrians. The IR-UWB radar transmits the pulse signal and measures the distance to a target using the time difference in the reflected signal. The radar signal gets reflected at each $1$ ns, and the Electromagnetic waves (EM waves) travel a $30$ cm distance. Hence, we get reflected signals at $15$ cm at every $1$ ns. The time instances when we receive the reflected signal are known as taps. In the proposed system, IR-UWB has a field of view of $135^\circ$, and the total taps is $56$, in which UASW uses $15$ taps, resulting in a range of $2.25$ meters.
%\textcolor{blue}{FOV is 135 degree. Tap: after transmitting the reference signal, Rx antennas will receive reflected signals at each nanosecond (1 ns). In 1 ns, EM waves travel 30 cm. So we get reflected signals at 15 cm at every 1 ns. This is known as taps. Our HW has 56 taps, in which UASW uses 15 taps (so 15x15 cm = 2.25 meters). Ranging Interval: The interval at which the received signals are transferred from UWB HW to the application. }

\subsection{System Architecture}
\label{sec:software}
\begin{figure}
	\centering
	\includegraphics[scale=0.66]{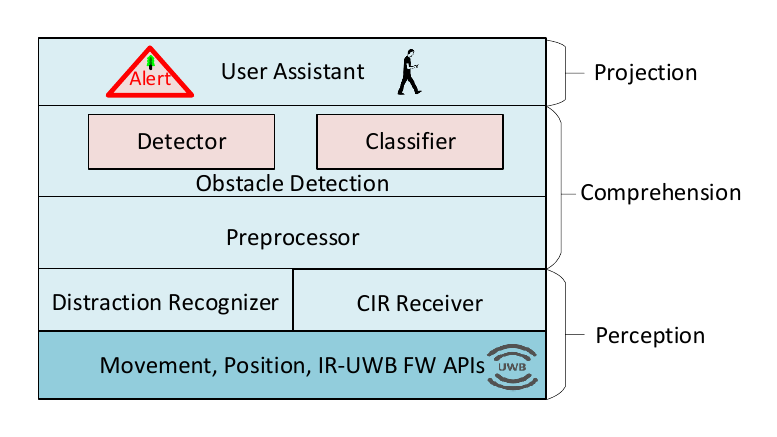}
 \vspace{-0.3cm}
	\caption{Architecture of UWB-assisted Safe Walk (UASW)}
	\label{fig:architecture}
 \vspace{-0.5cm}
\end{figure}
%Here, we discuss the key modules integrated into the proposed UASW system. 
Figure \ref{fig:architecture} shows UASW architecture, which encompasses three main steps in the SA process: \textit{perception}, \textit{comprehension}, and \textit{projection for decision-making}. The UASW system consists of five following modules.

\begin{itemize}
    \item \textbf{Distraction Recognizer:} This module is responsible for identifying distracted walking by utilizing a combination of sensor data and information about phone usage as triggers for the UASW system. 
    \vspace{0.1cm}
    \item \textbf{CIR Receiver:} Once distracted walking is detected, the CIR Receiver collects its CIR data in a complex baseband format. The magnitude of the CIR taps can be interpreted as the strength of the reflected signal, where a higher value indicates the presence of a reflecting object. % at the tap.
    %\vspace{0.1cm}
    \item \textbf{Preprocessor:} It parses each CIR sample, consisting of $56$ tap indices represented in hex-encoded complex numbers. These tap indices correspond to the range bin index from $B_{0:55} = 0:55$. Let the distance traveled at each tap be $D_{0:55}$. Each pulse radar tap interval is considered as $1$ ns apart. The distance of the first tap from the device, known as the tap length $D_{1}$ is given as
    \begin{equation}
     D_{1} = \frac{(B_1 \times \eta \times c)}{2}, 
    \end{equation}
    where $\eta$ is the time interval between taps ($1$ ns) and $c$ is the speed of light at which electromagnetic waves travel. The distance corresponding to $B_1$ range index is given as $D_{1} = 15 \pm 7.5$ cm and for $B_2$, the distance $D_2$ is given as $D_{2} = 30 \pm 7.5$ cm, and so on.   
    %For $B_1 = 1$, the range bin index $B_{1}$ is at a distance of $D_{1} = 15 \pm 7.5$ cm, $B_{2}$ is at a distance of $D_{2} = 30 \pm 7.5$ cm, and so on. %-- Not clear
    %\textcolor{blue}{After transmitting the reference signal, Rx antennas will receive reflected signals at each nanosecond (1 ns). In 1 ns, EM waves travel 30 cm. So we get reflected signals at 15 cm at every 1 ns. This is known as taps. Our HW has a total of 56 taps, in which UASW uses 15 taps (so 15x15 cm = 2.25 meters). Here, each tap is represented at $B_1$ $B_2$. Also, The distance it represents in $D_1$ $D_2$ } 
    In addition, the Preprocessor calibrates the radar distance to determine the range bin index corresponding to zero distance ($B_{0}$) from the CIR data. Once this calibration is performed, the Preprocessor extracts features from the subsequent taps using Fourier transformation.
    \vspace{0.1cm}
    \item \textbf{Obstacle Detection:} This module consists of two sub-modules that operate sequentially: the \textit{Detector} and the \textit{Classifier}. The \textit{Detector} uses a rule-based algorithm to identify obstacles, and the \textit{Classifier} employs various machine learning algorithms to categorize these identified obstacles according to attributes such as material type (such as glass, concrete, and human), wetness or dryness, and whether they are living or non-living entities.
    %The \textit{detector} utilizes a rule-based algorithm to detect obstacles, while the \textit{classifier} applies a range of ML algorithms to classify the detected obstacles as surfaces based on their material type (such as glass, concrete, etc.), wetness or dryness, and living or non-living objects.
   % \vspace{0.1cm}
    \item \textbf{User Assistant:} This module includes an application that serves as an alert and assistance provider, delivering real-time notifications regarding any modifications in the terrain or obstacles along the path. Furthermore, the severity of the alerts varies depending on the type and characteristics of the floor and the encountered obstacle.
\end{itemize}

\subsection{Operational Overview}
\begin{figure}
	\centering
	\includegraphics[scale=0.42]{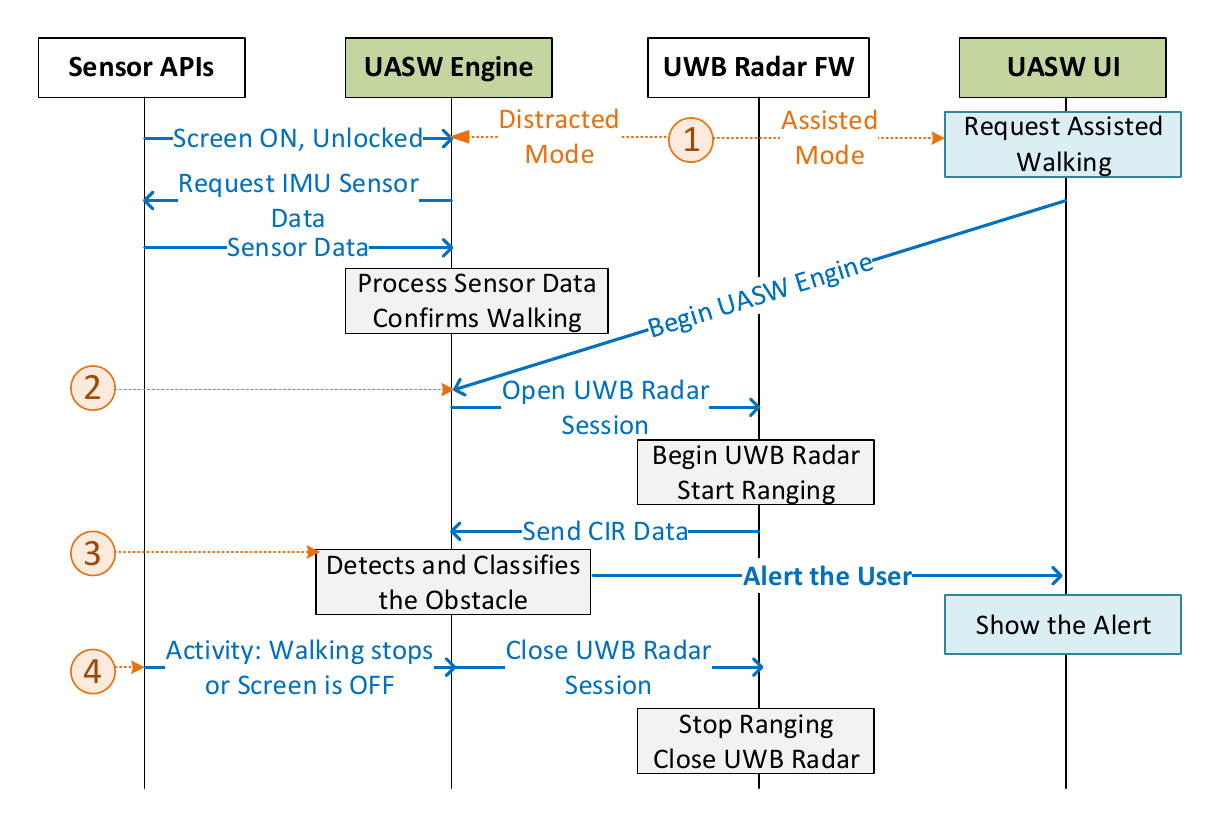}
	\caption{Operation flow for distracted walking and assisted walking modes}
	\label{flow}
\end{figure}

Figure \ref{flow} shows the stepwise operational overview of the UASW. There are two modes for enabling UASW: (i) \textit{Distracted Walking}, and (ii) \textit{Assisted Walking}. Both modes utilize similar preprocessing and obstacle detection mechanisms, differing only in the activation procedures at step (1) from the user's perspective. For \textit{Distracted Walking} mode, UASW employs Activity Recognition APIs \cite{actreg} and sensor data to verify that the user is walking while using a smartphone. The screen state detects smartphone usage, and the UASW is enabled if the user is detected as walking. For \textit{Assisted Walking}, the user requests to enable assisted walking through the UASW application on purpose. In step (2), UASW initiates an IR-UWB radar session to capture the CIRs. Thereafter, in step (3), the Preprocessor prepares the CIRs for Obstacle Detection, identifying and classifying obstacles to optimize the alerts. Once an obstacle and its type are confirmed, UASW generates notifications to alert the user. Finally, in step (4), the application instructs the IR-UWB radar to terminate the session if the user stops walking or using the phone.%, UASW instructs the IR-UWB radar to terminate the session. 

% In addition to its primary function of detecting distracted walking and providing timely alerts, UASW also includes an assisted walking mode to aid individuals with mobility challenges. UASW operates in two modes:

\begin{figure*}
	\centering
	\includegraphics[scale=0.55]{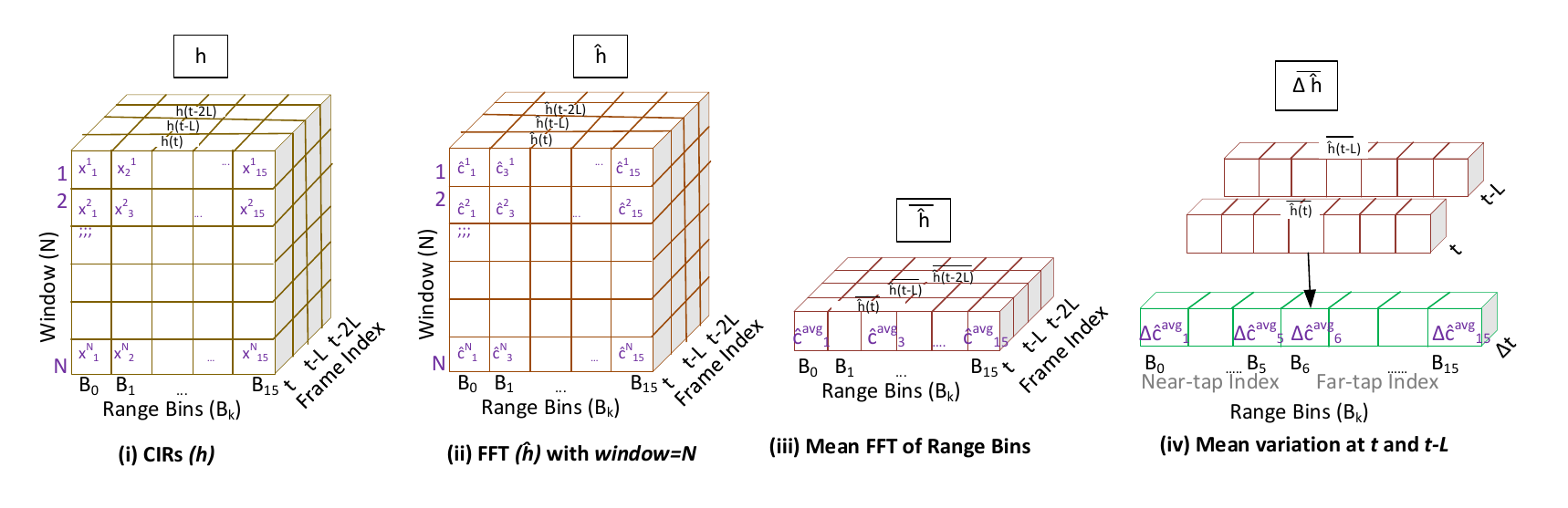}
	\caption{CIR preprocessing includes (i) Windowing the recent CIRs into a 3-dimensional matrix of height = $N$ (CIRs in CPI), width = $15$ (Range Bins), and depth = $4$ Frames. Frames are created with a sliding window of size $L$ ($N>L$).  (ii) Performing FFT for each frame in the buffer. (iii) Calculating mean FFT over CPI for each frame and  (iv) Finding the mean difference with Frame index $t$ and $t-L$.}
	\vspace{-5pt}
	\label{fig:fig_cir}
\end{figure*}

%\textbf{Assisted Walking mode}: To assist visually impaired users and provide awareness in low-visibility areas, UASW offers an Assisted Walking mode. As shown in Figure \ref{flow2}, (1) the user can request to enable Assisted Walking through the UASW application. (2) Upon receiving the request, the UASW engine initiates an IR-UWB radar session to receive data. The obstacle detector and classifier operate similarly to the Distracted Walking mode. (3) Subsequently, alerts are generated and delivered to the user. (4) The user can disable Assisted Walking once assistance is no longer required.
%\vspace{-1.7em}

\section{Methodology}
\label{sec:methodology}
%The methodology section presents the mechanisms for preprocessing the CIR data obtained from the IR-UWB radar and the techniques used for obstacle detection and classification. Furthermore, the integration of machine learning enables the system to adapt and enhance its performance over time.
%\tetxbf{CPI what is CPI}. %For a given vector $h = h_0, h_1, ..., h_{N-1}$ and $k = 0, 1, ..., N-1$, the FFT can be expressed as follows \cite{fft}:
%\begin{equation}
%\hat{h_k} = \frac{1}{\sqrt{N}}\sum_{n=0}^{N-1}h_n \times \phi^{nk}_N,
%\end{equation}
%where $\phi_N = \exp{-2\pi i/N}$. 

%\begin{equation}
%\overline{\hat{h}} = \frac{\sum_{n=0}^{N-1}\hat{h_n}}{N}.
%\end{equation}
\subsection{Obstacle Detection}
\label{detection}
We apply the Fast Fourier Transform (FFT) for feature extraction on time-based CIR data. The FFT window size is $N$, and the mean of frequency domain data ($\hat{h}_k$) is computed over the coherent processing interval (CPI) for each ranging interval. %(\textbf{what is ranging interval?}\textcolor{blue}{The CIR from UWB HW is transferred to the application layer in batches. It is configured by the value called Ranging Interval. So we get CIRs for every 40 ms}) 
 UASW employs a rule-based approach for obstacle detection which utilizes the mean value of $\overline{\hat{h}}$ for the range bins $B_1$ to $B_{15}$ as input features representing the distance from $15$ cm to $2.25$ meters.% , (\textbf{need to mention why we are choosing B1 to B15} \textcolor{blue}{As I mentioned earlier it represents the distance from 1x15 to 15x15, that is close to 2.25 meters}).

\textbf{Rule-based approach:} In the rule-based approach, UASW analyzes the mean variation of the FFT within the CPI (denoted as $\Delta\overline{\hat{h}}(k, k-1)$) and any significant change in mean value helps in obstacle detection. Based on the current Rx gain index, a threshold $\gamma$ is set as $20$. The algorithm determines the presence of obstacles based on the divergence between the close tap indexes ($B_1$ to $B_3$) and the far tap indexes ($B_4$ to $B_{15}$). The algorithm detects objects ahead of the user by examining the mean variation of three consecutive far-tap indexes. Additionally, false positive detection is addressed through the mean difference of the near-tap indexes. The presence of an obstacle is determined by the following function $f(\hat{h})$

\begin{equation}
\label{eq1}
f(\hat{h}) = 
\begin{cases}
1 & \text{if $\sigma \leq 2$ and $\Delta\overline{\hat{h}}_{(i, i+1, i+2)} > \gamma$}, \\ 
& \text{where, }  i \in \{B_4,.., B_{13}\}  \\
0 & \text{otherwise}
\end{cases}
\end{equation}

Here, $\gamma$ represents the mean variation threshold for the current Rx Gain, and $\sigma$ is the false positive detection function. $\sigma$ is updated according to the following equation
\begin{equation}
\sigma = \sigma + 1 ~\forall~ \Delta\overline{\hat{h}}_l > \gamma, \text{where, }l \in \{B_1,.., B_{3}\}.
\end{equation}
\vspace{-0.5cm}
%The $\sigma$ is updated 

%\textbf{ML-based Approach:}
%\textcolor{blue}{This session is not so clear. The idea was to tell that Obstacle Detection can also be tried with ML. I think we can tell that Detection is completely SP method. So we can remove this session as well.}
%We employ ML algorithms for obstacle detection where $\overline{\hat{h}}$ for range bins $B_1$ to $B_{15}$ are the input features. The dataset was collected from indoor office premises, sidewalks, and crowded places. Objects within a range of 1 to 1.5 meters from the smartphone user are labelled as $label=1$, while the rest are labelled as $label=0$. The ML pipeline standardized the input features and applied stratified sampling for training. Hyperparameter tuning was performed using the grid search for each model. \textbf{please mention the training dataset, test dataset details, which ML algorithms are being implemented, and what activation function is used}
\subsection{Obstacle Classification}
Once an obstacle is detected ahead of a distracted smartphone user, UASW provides additional details such as obstacle type, surface nature, and movement. The \textit{classifier} performs inference only when the \textit{detector} confirms the presence of an obstacle. The training dataset consists of 50,000 entries and includes obstacle types such as glass, concrete, wood, and humans. Labels are also assigned for surface type (dry or wet) and movement type (static or mobile). The input features undergo preprocessing steps, including standardization, imputation, and stratified sampling, similar to the detection model. We compared various classification models such as logistic regression, decision tree, random forest, K neared neighbor (KNN) classifier, stochastic gradient descent (SGD) classifier, gradient boosting, support vector machine classifier (SVC), and artificial neural network (ANN) architecture consisting of multiple dense layers. The output layer of the network is configured to predict class labels for the input data using softmax activation. Model training involved optimizing the parameters using the categorical cross-entropy loss function, and Adam optimizer.% is used for (\textbf{fow what??} \textcolor{blue}{remove 'is used for'}).

\section{Performance Analysis}
\label{sec:perfm_analysis}

	\begin{table}
		\centering
		\caption{IR-UWB Radar Configuration}
		\label{configs}
		\begin{tabular}{|l|l|}
			\hline
			\textbf{PARAMETER} & \textbf{VALUE} \\ \hline
			Tx Antenna & 1 \\ \hline
			Rx Antenna & 1 \\ \hline
			Pulse Repetition Frequency (PRF) & 200 Hz \\ \hline
			Radar Frame Repetition Interval (RFRI) & 5 ms \\ \hline
			Ranging Interval (RI) & 40 ms \\ \hline
			Tap Interval ($\eta$) & 1 ns \\ \hline
			Tap Distance ($D_i$) & ~15 cm \\ \hline
			Coherent Processing Interval (CPI) & 320 ms \\ \hline
			No. of CIRs in CPI ($N$) & 64 \\ \hline
			Frame Length ($L$) & 40 ms \\ \hline
		\end{tabular}
  \vspace{-0.5cm}
	\end{table} 
\subsection{Parameter Configuration}
\label{config}
The Device Under Test (DUT) contains a UWB chipset with radar functionality. We implemented UASW on Samsung devices running on Android 13 and equipped with an IR-UWB radar. Note that the model and chipset details are withheld for confidentiality.  %Table \ref{tab:radar_confg} shows the hardware configuration of the radar and the parameters used for the experiments.     
This DUT includes a transmission (Tx) antenna for signal transmission and three reception (Rx) antennas for capturing the reflected signal, known as the CIR. Currently, UASW utilizes one Rx antenna and one Tx antenna from the chipset. Table \ref{configs} summarizes the configuration details of the IR-UWB chipset used. We perform a radar distance calibration to detect the position of the $0^{\text{th}}$ range bin ($B_0$) and subsequently identify Tap $0$ to Tap $15$ ($B_0$ to $B_{15}$). %A CPI of $320$ ms was chosen, considering the desired accuracy and power usage. With an RFRI (Radar Frame Repetition Interval) of $5$ ms, the number of CIRs captured during each CPI is denoted as $N = 64$. 
Furthermore, a sliding window of length $L = 40$ ms is applied for CIR windowing to ensure the appropriate segmentation and processing of the CIR data within the defined time frame. Figure \ref{fig:fig_cir} shows the CIR processing for obstacle detection and classification. Initially, UASW configures the CPI, representing the total time to sample a group of multiple CIRs with the same PRF and frequency. The recent CIRs are windowed into a 3-dimensional matrix with a height of $N$ (CIRs in CPI), a width of 15 (range bins), and a depth of $4$ frames. The frames are created with a sliding window of duration $L$. 

\begin{figure}
	\centering
	%\vspace{-5pt}
	\includegraphics[scale=0.34]{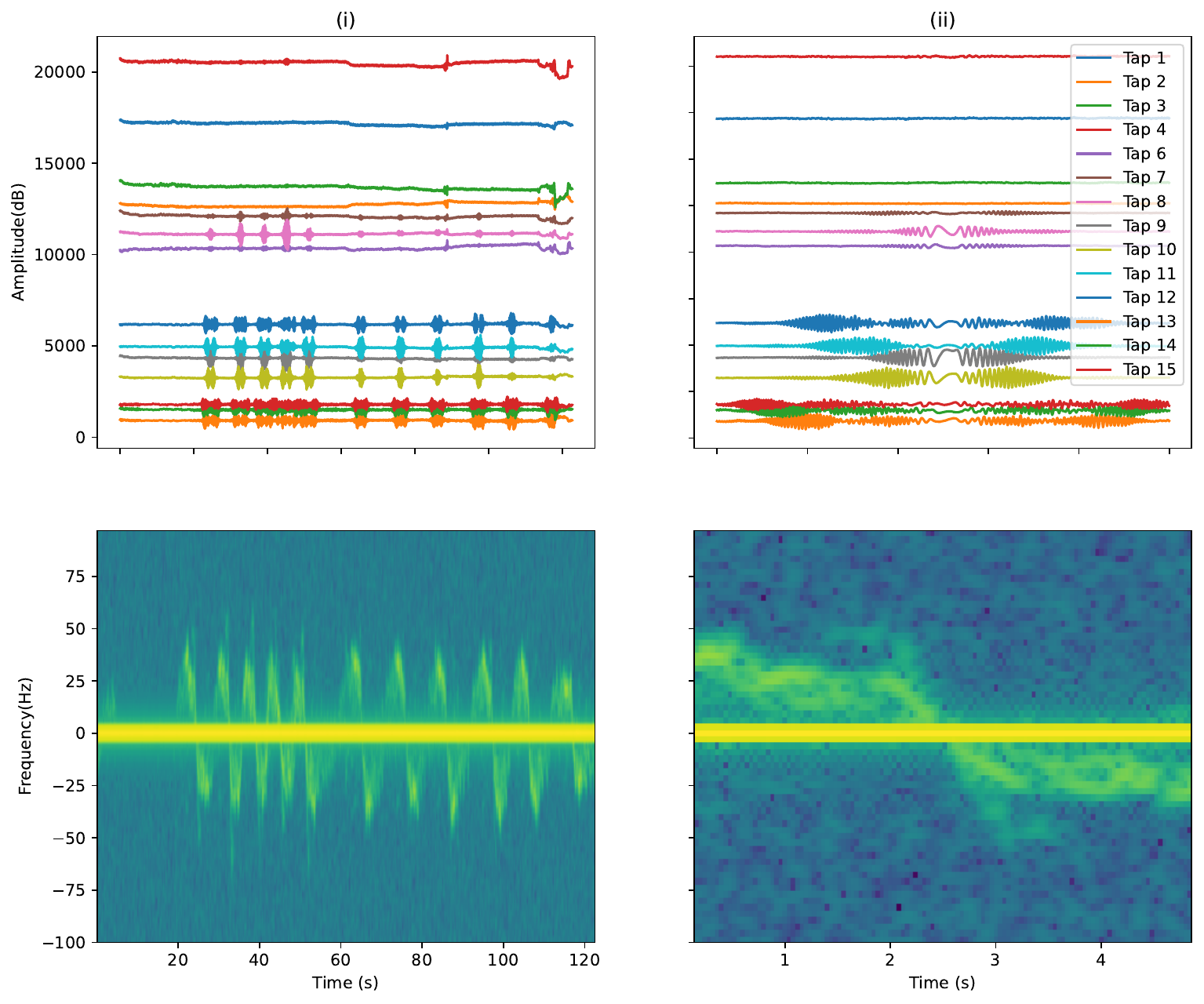}
	%\vspace{-25pt}
	\caption{CIR Variation for Tap $1$ to Tap $15$ and corresponding FFT spectrogram for Tap $15$ with $FFT_{window} = N = 64$ and $Overlap = N-L = 56$. (i) A distracted walking session for 120 seconds. (ii) When the user walks towards a glass door and moves back.}
	\vspace{-7pt}
	\label{fig:cir}
\end{figure}
\subsection{Data Collection}
The UASW system is developed as a dedicated application for Android 13 smartphones. Multiple users are assigned to walk with their smartphones in hand across various scenarios to gather comprehensive data. Extensive experiments are conducted in different environments, including indoor locations, sidewalks, and crowded common areas within our office premises. We developed a data logging application with a manual obstacle-marking feature to facilitate data collection. This feature allows users to annotate the type and movement of encountered obstacles. Users collected over 50,000 obstacle data points, providing valuable information on the obstacles' material, surface type, and movement characteristics. %The rule-based obstacle detection algorithm is also employed to ensure manual labelling accuracy.
The data cleaning process and imputation are performed to enhance the quality of collected data. Figure \ref{fig:cir} illustrates the CIR fluctuations caused by obstacles and corresponding FFT spectrogram generated using matplotlib library in Python\cite{matplotlib}. The variations in spectrograms are the foundation for the obstacle detection and classification methods, utilizing the CIR information to identify and categorize objects encountered during the experiments.

%\subsection{Evaluation Metrics}
%The UASW experiments focused on response time and accuracy metrics to assess the system's performance. Response time refers to the speed at which the SAS detects and alerts users to potential environmental hazards or obstacles. A shorter response time is desirable as it allows users to react promptly and mitigate risks. Accuracy pertains to the reliability and precision of the SAS in accurately identifying and classifying obstacles. High accuracy ensures that users receive relevant and trustworthy alerts, minimizing false positives or negatives. By prioritizing these critical metrics, the experiments provided valuable insights into the performance and capabilities of UASW.

\subsection{Power Consumption}
UASW system effectively minimizes the active time of the IR-UWB radar session on the Android device to address power consumption concerns. By leveraging the capability of detecting distracted walking, UASW initiates the radar session only when necessary, reducing the overall active duration. Additionally, UASW implements a radar session cut-off mechanism, which automatically terminates the radar ranging after a specific time interval. The radar session is set to cease after $10$ seconds, with a cut-off threshold of $2$ seconds. This logic ensures that the radar remains active reasonably while avoiding unnecessary power consumption during extended periods. %By implementing these power-saving measures, the UASW system achieves notable reductions in power consumption. 
The IR-UWB radar chipset typically consumes approximately $39.8$ mA for an RFRI of $5$ ms, whereas UASW consumes $24.7$ mA, resulting in significant power savings.
%%%%%%%%%%%%%%%%%%%%%%%%%%%%%%%%%%%%%%%%%%%%%%
%%%%%%%%%%%%%%%%%%%%%%%%%%%%%%%%%%%%%%%%%%%%%%
\begin{table}
    \centering
		\caption{The total response time of UASW system.}
		\label{times}
		\begin{tabular}{|l|p{2.1cm}|p{2cm}|}
			\hline
			\textbf{Process}             & \textbf{No Ensemble} & \textbf{With Ensemble} \\ \hline
			CIR Acquisition              & 5 ms to 40 ms (22.5 ms)          & 85 to 120 ms (102.5 ms)          \\ \hline
			Pre-processing           & 0.8 ms                 & 2.4 ms                    \\ \hline
			Detection           & 1.1 ms                 & 3.3 ms                    \\ \hline
			Classification      & 2.4 ms                 & 7.2 ms                    \\ \hline
			\textbf{Response Time} & \textbf{26.8 ms}      & \textbf{115.4 ms}       \\ \hline
		\end{tabular}
	\end{table}
\begin{figure}
    \centering
	%\vspace{-5pt}
    \includegraphics[scale=0.5]{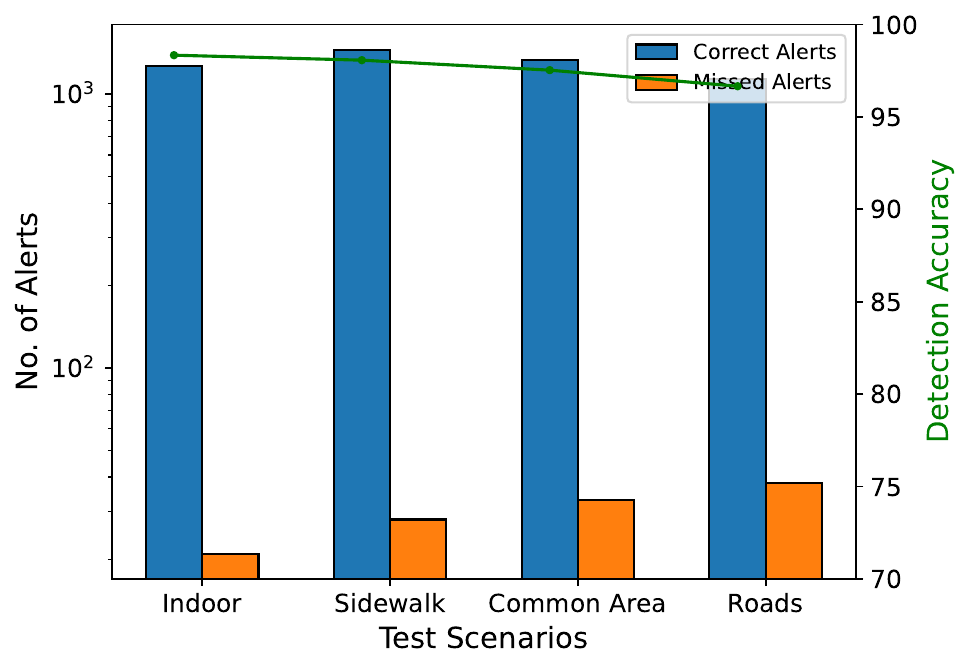}
	\vspace{-7pt}
    \caption{Detection accuracy of proposed algorithm for different test scenarios}
	\vspace{-10pt}
	\label{fig:det}
\end{figure}
\section{Results and Discussions}
\label{sec:results}
In the context of SAS, response time and accuracy are crucial metrics for evaluating its effectiveness. Response time refers to the speed at which the SAS detects and alerts users, and accuracy pertains to reliability and precision in accurately identifying and classifying obstacles.
The average response time of the UASW system includes two components: (i) Time taken for acquiring the CIRs from radar and (ii) Time taken for UASW logic to execute. Considering the RFRI is $5$ ms for a ranging interval of $40$ ms, the total time for CIR acquisition varies from $85$ to $120$ ms and $5$ ms to $40$ ms for classification with and without the ensembling method. CIR preprocessing steps involve feature extraction using FFT, which takes up to $0.8$ ms. Similarly, the response time for obstacle detection is measured at $1.1$ ms, while obstacle classification requires $2.4$ ms. As listed in Table.\ref{times}, the total response time ranged from $26.8$ ms to $115.4$ ms, providing real-time feedback to users. Table \ref{times} shows that the proposed system is significantly fast even while including ensembling.

\begin{figure}
	\centering
	%\vspace{-5pt}
	\includegraphics[scale=0.5]{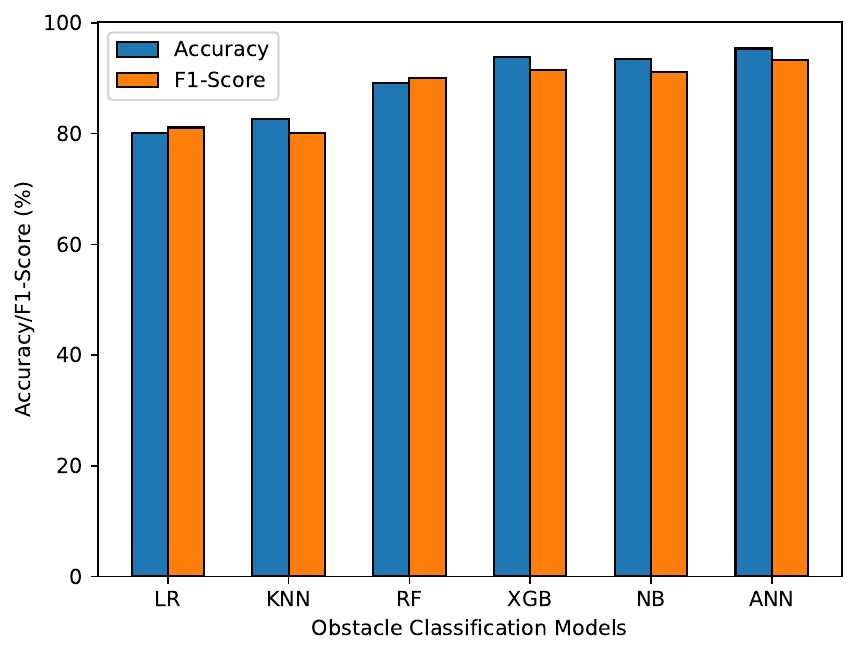}
	%\vspace{-25pt}
	\caption{Accuracy and F1-Score of various obstacle classification models}
	%\vspace{-12pt}
	\label{fig:class}
\end{figure}

\begin{figure}[h!]
	\centering
	%\vspace{-5pt}
	\includegraphics[scale=0.31]{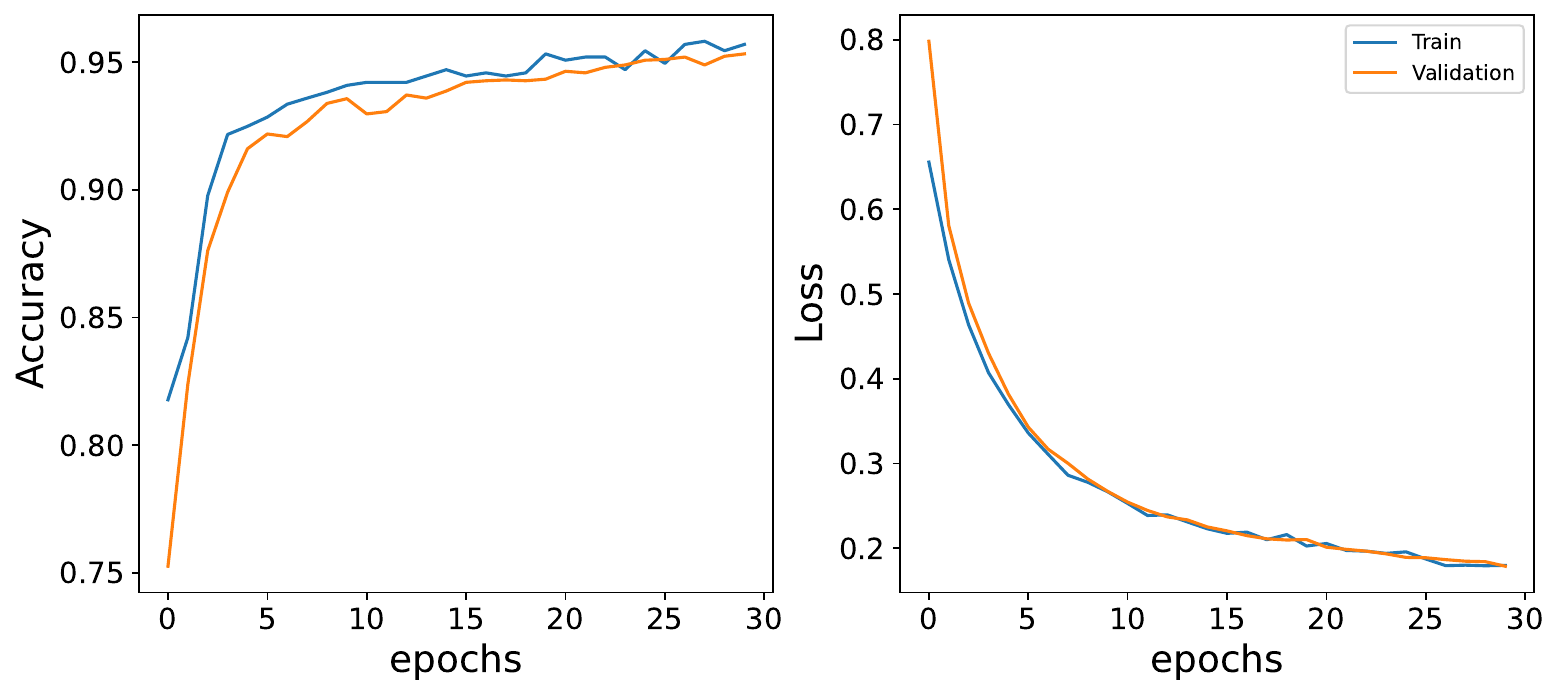}
	%\vspace{-10pt}
	\caption{Accuracy and Loss plot for ANN-based obstacle classification model}
	\vspace{-0.2cm}
	\label{fig_nn}
\end{figure}

The accuracy of the rule-based obstacle detection algorithm is evaluated across four different scenarios. Over several days of 60-minute sessions, the algorithm exhibited an accuracy of $97.38$\% (averaged across four scenarios) with a minimal miss-hit ratio of $0.025$\%, as shown in Figure \ref{fig:det}. The performance of various ML and ANN models for the obstacle classification task is shown in Figure \ref{fig:class}. Additionally, Table \ref{tab:annmodel} lists the various configurations of the ANN models. We deployed an ANN model on an Android smartphone, which achieved an impressive accuracy of $95.38$\% as shown in Figure \ref{fig_nn}. Importantly, we measured the worst-case on-device inference delay at $2.4$ ms using the TensorFlow Lite library for Android, ensuring efficient and timely obstacle classification. The proposed UASW system demonstrates promising results in obstacle detection and classification. 
% \begin{table}[h!]
% \caption{Accuracy for models with and without feature engineering}
% \label{tab:mlmodel}
% \begin{tabular}{|l|l|l|}
% \hline
% Model               & Accuracy (With FE) & Accuracy (Without FE) \\\hline
% Logistic Regression & 80.19              & 80.01                 \\ \hline
% KN Classifier       & 82.6               & 81.05                 \\\hline
% Random Forest       & 89.14              & 87.15                 \\\hline
% XG Boost            & 93.8               & 94.87                 \\\hline
% Naive Bayes         & 93.56              & 93.56              \\  \hline
% \end{tabular}
% \end{table}

\begin{table}
\centering
\caption{Experimental results for various ANN configurations}
\label{tab:annmodel}
\begin{tabular}{|l|l|l|}
\hline
\textbf{NN Layers} & \textbf{F1 score} & \textbf{Accuracy (\%)} \\ \hline
Layers: 1   & 86.34 & 89.2 \\  \hline
Layers: 2, Neurons: 12 & \textbf{93.22}& \textbf{95.38}  \\ \hline 
Layers: 2, Neurons: 24 & 90.5& 94.2  \\ \hline 
Layers: 3, Neurons: 18 &  91.23  & 93.89  \\ \hline
\end{tabular}
\end{table}
We also compare the UASW solution with BLE-based (similar to \cite{bledw}) and Camera-based obstacle detection \cite{vision1} solutions for distracted pedestrians.
The BLE Beacon-Based solution requires the installation of BLE beacons throughout the road infrastructure, leading to significant installation overhead due to the widespread deployment of beacons.
Additionally, the communication latency between the beacons and smartphones can be relatively high, as illustrated in Table \ref{tab:comp}. The camera-based approach utilizes smartphone cameras for obstacle detection. However, this method often consumes substantial power and raises privacy concerns as it involves continuous image capture and processing. In contrast, UASW leverages IR-UWB radar connectivity for obstacle detection. It offers a low-power and low-latency solution that minimizes the impact on smartphone battery life and delivers real-time obstacle detection without having any privacy concerns.

\begin{figure}[t!]
	\centering
	%\vspace{-5pt}
	\includegraphics[scale=0.36]{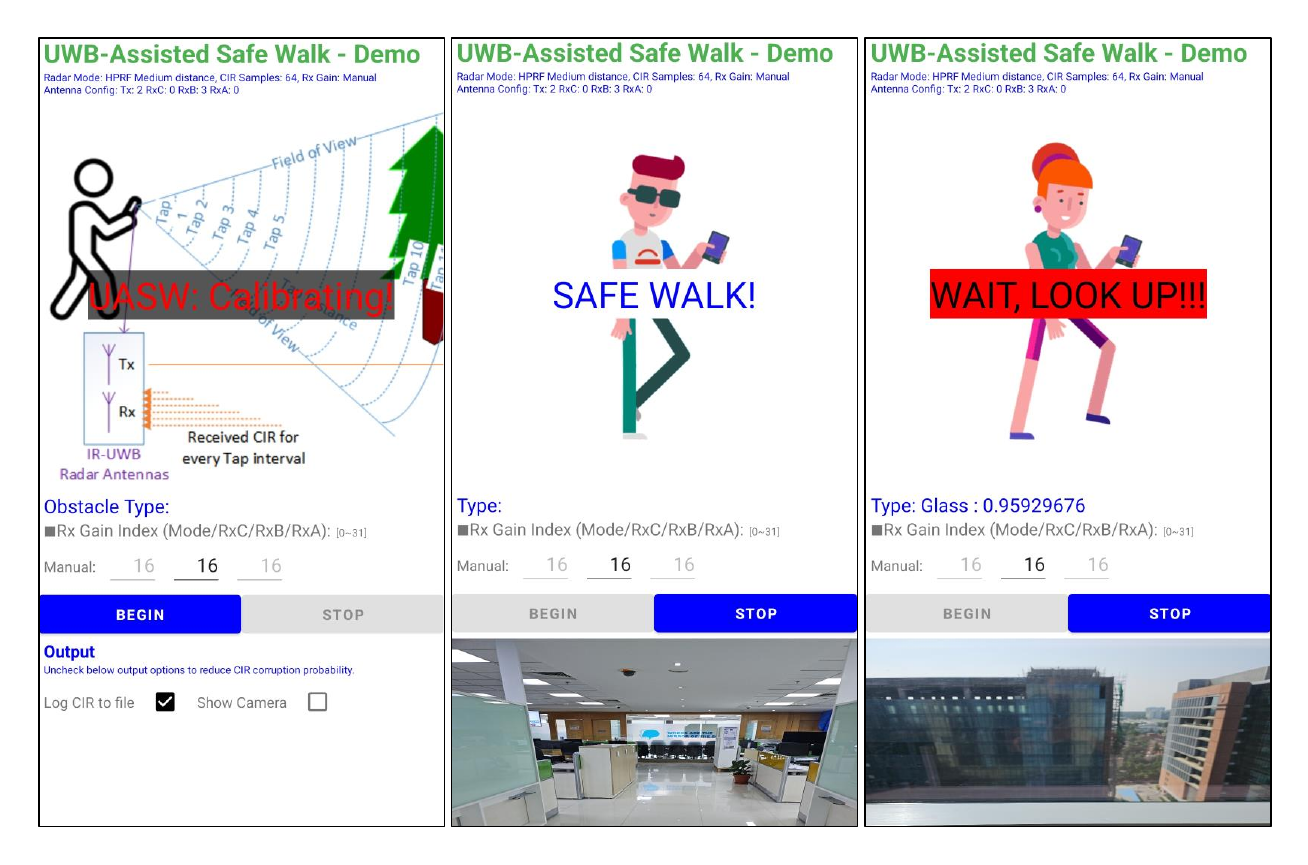}
	%\vspace{-10pt}
	\caption{Screenshots from the UASW Demo Application during (i) calibration, (ii) no obstacles present, and (iii) the obstacle is present. The camera preview is added for demo purposes. Obstacle type and classification confidence scores are also displayed.}
	%\vspace{-12pt}
	\label{fig:fig_app}
\end{figure}

Figure \ref{fig:fig_app} shows the snapshots during the application usage with DUT. The results highlight the efficacy and potential use case of the proposed UASW system in accurately detecting and classifying obstacles, thereby enhancing situational awareness for smartphone-distracted pedestrians. The UASW system has low response time and high accuracy, which contributes to its reliability in providing timely alerts and relevant information to users, enabling them to navigate their surroundings safely.
\begin{table}[t!]
\caption{Comparison of UASW with BLE and Vision-based Solutions}
\label{tab:comp}
\begin{tabular}{|l|l|l|l|}
\hline
\textbf{Solution} & \textbf{Power Usage} & \textbf{Latency} & \textbf{Install. Overhead} \\ \hline
\textbf{UASW}     & $24.7$ mA                    & $26.8$ ms          & NA                         \\ \hline
BLE-based         & $\sim20$ mA                & $200$ ms           & Yes                        \\ \hline
Vision-based      & $\sim1050$ mA              & $\sim40$ ms      & NA                         \\ \hline
\end{tabular}
\end{table}

\section{Conclusion}
\label{sec:conclusion}
This paper proposes a novel, technologically advanced solution to a significant growing pedestrian safety problem. 
%This paper focuses on developing a real-time situational awareness system for personal safety. 
The proposed solution leverages IR-UWB radar technology, available in modern smartphones, creating a cost-effective and convenient solution to situational awareness. We also demonstrate two ways the proposed system can be utilized: distracted walking mode and assisted walking mode, which aids individuals with mobility challenges.
The outcomes substantiate the efficacy of the proposed system, boasting an impressive 95.38\% accuracy for object classification and 97.63 \% accuracy for object detection. The results show a satisfactory inference delay of $26.8$ ms. In the future, it is proposed to consider more challenging on-road scenarios and incorporate the direction-finding algorithms to improve the performance of the UASW system further. 
\bibliographystyle{IEEEtran}
\bibliography{IEEEabrv,UASW}

%\section{Modified Results}

%The dataset contains the following classes. The classification takes place after an obstacle is detected. So entire data points are whenever an obstacle is ahead of the user.

% %1. Wall
% 2. Glass Wall
% 3. Human

% Also for terrain, when the phone is inclined towards the floor, we classify the characteristics in the following way

% 1. Flat vs. Stairs
% 2. Dry vs. Wet 

% Two separate classifications.

\end{document}